\documentclass[10pt,letterpaper]{article}

\usepackage{amsthm,amsmath}

\usepackage[utf8]{inputenc} 

\usepackage{nameref,hyperref}

\usepackage{graphicx}
\usepackage{xcolor}
\usepackage{multirow}
\usepackage{xspace}
\usepackage{bold-extra}

\usepackage{tikz}
\usetikzlibrary{shapes}
\usetikzlibrary{shapes.multipart}
\usetikzlibrary{fit}

\usepackage{amssymb}
\usepackage{mathrsfs}

\usepackage{geometry}
\usepackage{pdflscape}

\usepackage{booktabs}
\usepackage{makecell}

\usepackage{todonotes}

\newcommand{\mtt}[1]{\text{\texttt{#1}}}

\newcommand{\ie}{\emph{i.e.},\xspace}
\newcommand{\eg}{\emph{e.g.},\xspace}

\def\it#1{\text{\textit{#1}}}

\begin{document}
\vspace*{0.35in}

\begin{flushleft}
{\Large
\textbf\newline{Investigating ADR mechanisms with\\
knowledge graph mining and\\
explainable AI}
}
\smallskip

Emmanuel Bresso\textsuperscript{1,$\diamond$}, 
Pierre Monnin\textsuperscript{1,2,$\diamond$},
C\'edric Bousquet\textsuperscript{3,4},
François-Elie Calvier\textsuperscript{4},\\
Ndeye-Coumba Ndiaye\textsuperscript{5},
Nadine Petitpain\textsuperscript{6},
Malika Sma\"il-Tabbone\textsuperscript{1},
Adrien Coulet\textsuperscript{1,7,$\star$}\\

\bigskip

\text{\textsuperscript{1}} Université de Lorraine, CNRS, Inria, LORIA, Nancy, France
\\
\text{\textsuperscript{2}} Orange Labs, Belfort, France
\\
\text{\textsuperscript{3}} LIMICS, Inserm UMR1142, Université Sorbonne Paris Nord, Sorbonne Université, Paris, France
\\
\text{\textsuperscript{4}} CHU de St Etienne, St Etienne, France
\\
\text{\textsuperscript{5}} NGERE, Inserm U1256, Université de Lorraine, Nancy, France 
\\
\text{\textsuperscript{6}} Centre Régional de Pharmacovigilance, CHRU de Nancy, Nancy, France 
\\
\text{\textsuperscript{7}} Inria Paris, Inserm UMR1138, Université de Paris, Sorbonne Université, Paris, France
\\
\bigskip
\textsuperscript{$\diamond$} These authors contributed equally to this work.\\
\textsuperscript{$\star$} corresponding author: \texttt{adrien.coulet@inria.fr}

\end{flushleft}

\section*{Abstract}
\textbf{Background} 
Adverse Drug Reactions (ADRs) are statistically characterized within randomized clinical trials and postmarketing pharmacovigilance, but their molecular mechanism remains unknown in most cases. This is true even for hepatic or skin toxicities, which are classically monitored during drug design. Aside from clinical trials, many elements of knowledge about drug ingredients are available in open-access knowledge graphs, such as their properties, interactions, or involvements in pathways. In addition, drug classifications that label drugs as either causative or not for several ADRs, have been established.
\textbf{Methods} 
We propose in this paper to mine knowledge graphs for identifying biomolecular features that may enable reproducing automatically expert classifications that distinguish drug causative or not for a given type of ADR. 
In an explainable AI perspective, we explore simple classification techniques such as Decision Trees and Classification Rules because they provide human-readable models, which explain the classification itself, but may also provide elements of explanation for molecular mechanisms behind ADRs.
In summary, \textit{(i)} we mine a knowledge graph for features, \textit{(ii)} we train classifiers at distinguishing, on the basis of extracted features; drugs associated or not with two commonly monitored ADRs: drug-induced liver injuries (DILI) and severe cutaneous adverse reactions (SCAR);  \textit{(iii)} we isolate features that are both efficient in reproducing expert classifications and interpretable by experts (\textit{i.e.}, Gene Ontology terms, drug targets, or pathway names); and \textit{(iv)} we manually evaluate how they may be explanatory. 
\textbf{Results} 
Extracted features reproduce with a good fidelity classifications of drugs causative or not for DILI and SCAR (Accuracy=\it{0.74} and \it{0.81}, respectively). 
Experts fully agreed that \it{73\%} and \it{38\%} of the most discriminative features are possibly explanatory for DILI and SCAR, respectively; and partially agreed (2/3) for \it{90\%} and \it{77\%} of them. 
\textbf{Conclusion}
Knowledge graphs provide sufficiently diverse features to enable simple and explainable models to distinguish between drugs that are causative or not for ADRs. In addition to explaining classifications, most discriminative features appear to be good candidates for investigating ADR mechanisms further.

\vspace{2em}

\noindent\textbf{Keywords:}
Adverse drug reaction, Molecular mechanism, Mechanism of action, Explanation, Data mining, Machine learning, Knowledge graph, Explainable AI

\section*{Background}
Molecular mechanisms behind harmful or beneficial effects of drugs are largely unknown.  
For instance, the molecular mechanism of highly prescribed drugs acetaminophen, lithium, and metformin is not completely understood.
Indeed, drug development process relies mainly on randomized clinical trials and postmarketing pharmacovigilance that evaluate drug efficacy and safety, independently from any mechanistic investigation \cite{ciociola2014}. 
However, understanding a drug's mechanism is fruitful: it can guide drug development, improve drug safety, and enable precision medicine, through better dosing or combination of drugs \cite{mechanism_matters2010}. 
Aside from this partial ignorance, many elements of knowledge about drug ingredients are available in open-access knowledge graphs, such as their chemical and physical properties, their interactions with biomolecules such as their targets, or their involvements in biological pathways or molecular functions \cite{Kamdar2019EnablingWD}. 
Knowledge graphs can be broadly defined as graphs of data with the intent to compose knowledge. 
Here, we consider knowledge graphs represented using Semantic Web technologies, including RDF (Resource Description Framework) and URI (Uniform Resource Identifier)~\cite{bonatti_et_al:DR:2019:10328, berners2001}.
In such knowledge graphs, nodes represent \textit{entities}, also named \textit{individuals}, of a domain (\eg acetaminophen), \textit{classes} of individuals (\eg analgesics), or \textit{literals} (\eg strings, dates, numbers).
Literals are purposely discarded in this study.
Nodes are connected by directed edges that are labeled with \textit{predicates} (\eg \texttt{transportedBy}). 

We propose in this article to leverage elements of knowledge about drugs that lie in biomedical knowledge graphs to investigate ADR molecular mechanisms. To this aim, we experiment with knowledge graph as an input to machine learning approaches (\ie methods of Artificial Intelligence, commonly denoted AI) that are natively explainable. Indeed, Explainable AI usually refers to research on methods that provide explanatory elements to results (\ie a classification) of sub-symbolic approaches (\eg ensembles or Deep Neural Networks) \cite{arrieta2020}. In a broadly manner, we consider symbolic approaches that provide models that are interpretable by humans, and investigate if features of these models may be explanatory for biomolecular processes involved in ADRs. 
We particularly consider a knowledge graph named PGxLOD, which encompasses and connects drug, pathway, and biomolecule data \cite{monninLHRTJNC19}; and two particular types of ADRs: drug-induced liver injuries (DILI) and severe cutaneous adverse reactions (SCAR). We choose these types of ADRs first because hepatic or skin toxicities are commonly monitored during drug developement, because of their importance in pharmacovigilance~\cite{trifiro2019}. Indeed hepatic and skin events are frequently caused by drugs, and they are severe enough to potentially lead to drug withdrawal in Phase IV. Second, it exists good quality \emph{expert classifications} that label sets of drugs as either causative or not for these types of ADRs \cite{chen16,regiscar_notoriety_list}. 
First, our work identifies biomolecular features from our knowledge graph that enable an automatic reproduction of expert classifications. In particular we mine the graph for neighbors of drugs, paths and path patterns (\ie paths composed of general classes) rooted by drugs and passing by at least one entity of the following types: pathway, gene/protein, Gene Ontology (GO) term or MeSH term. 
Second, we isolate both predictive and interpretative features hypothesizing that, in addition to be explanatory for the classification, those may also be explanatory for ADR mechanisms.  
To this second aim, we consider simple, but explanatory classification techniques, \ie Decision Tree and propositional rule learner over extracted features, because they provide human-readable models in the form of rules. Finally, we isolate features that are both predictive and interpretable, and ask human experts if they consider them as possibly explanatory for ADRs.

A first family of related works can be described as \textit{explanatory}, where known Drug-ADR associations, such as those found in SIDER, are used to highlight molecular mechanisms that may be impacted in ADRs. 
A second family of works is \textit{predictive}, where data about molecular mechanisms (\eg GO molecular processes or KEGG pathways) are associated with drugs and used as features to predict ADRs \cite{ho2016}. 
Boland  \textit{et al.} survey existing works for both predicting ADRs and elucidating their mechanisms; they interestingly list data and knowledge resources that may support these efforts \cite{boland2016}. 

In the explanatory family, 
Lee \textit{et al.} associate Sider side effects and GO biological processes through drugs, by the combination of a Drug-Side effect and a Drug-Biological process networks \cite{lee2011}. Highlighted relationships are obtained using statistical approaches (\ie enrichment and T-score) and evaluated in regards with co-occurrences in PubMed abstracts.
Wallach \textit{et al.} link drugs to proteins through molecular docking, then to pathways through the KEGG database, and used logicistic regression and feature selection approaches to select pathways most probably impacted in side effects \cite{wallach2010}.
Bresso \textit{et al.} group frequently associated drug reactions and propose elements of explanations of their grouping using Inductive Logic Programming~ \cite{bresso2013}. Also, Chen \textit{et al.} proposed a computational algorithm to infer Protein-ADR relationships from a network of protein-protein interactions, ADR-ADR similarities and known protein-ADR relations~\cite{chen2016_sr}.

In the predictive family, 
Bean \textit{et al.} build a network of drugs, targets, indications and ADRs to select features that are good predictors for ADRs in a logistic regression setting. 
PhLeGrA proposes an analytic method based on Hidden Conditional Random Fields to allow the calculation of the probability of drug reactions given a input drug and a knowledge graph of drugs, proteins, pathways and phenotypes~\cite{kamdarM17}. Similarly, Mu\~noz \textit{et al.} proposed a specific way to extract features from knowledge graphs for ADR prediction~\cite{munozNV19}. They show that several multi-label learning models perform well for this task. 
Our work is similar to some extent, however it uses simpler but explanatory classifiers, and goes a step further by identifying features, subsequently proposed as explanatory elements for ADRs.
 Indeed, we hypothesize that within the large set of considered features, those that are both highly predictive and associated with a good level of interpretablity may suggest to experts plausible elements of explanation. In Dalleau \textit{et al.} knowledge graph mining serves a completion perspective and aims at inferring links between drugs and genes \cite{dalleau2017}. 
All PhLeGrA, Mu\~noz \textit{et al.}, Dalleau \textit{et al.}, and the present work illustrate the interest of aggregating several LOD (linked open data) sets for knowledge discovery and data mining tasks, as discussed by Ristoski and Paulheim~\cite{ristoskiP16}. 

Shi and Weninger~\cite{shiW16} use a similar approach to ours, but from a fact checking perspective. 
Indeed, for each relation type $p$, a set $D^k_{(o_u,o_v)}$ of discriminative paths is learned. 
This set contains anchored predicate paths 
$o_u \xrightarrow[]{r_1} \xrightarrow[]{r_2} \dots \xrightarrow[]{r_{k}} o_v$ of length $k$ that describe a statement $o_u \xrightarrow[]{p} o_v$, where $o_v$ and $o_v$ are respectively the ontology classes associated with nodes $u$ and $v$.
To check whether a triple $s \xrightarrow[]{p} t$ is true, they use the learned set of discriminative paths for the relation $D^k_{(o_s,o_t)}$. In such sets, only paths with the most discriminative power are kept. Similarly to our approach, they use ontology class generalization but apply it only to start and end nodes $s$ and $t$ of the fact to be checked. This differs from our approach as we apply generalization on each intermediate node (see Methods Section for details). 
Additionally, their path modelling allows reverse traversal, \textit{i.e.} $\xrightarrow[]{p^{-1}}$ and constraints both source and target nodes, while we only constraint source nodes.
Previous works also use knowledge graph mining to provide explanations. Those include Explain-a-LOD that enriches statistical data sets with features from DBpedia. It uses correlation between attributes and rule learning to provide hypothesis explaining statistics~\cite{paulheim12}. Explain-a-LOD relies on FeGeLOD to extract features from the DBPedia knowledge graph~\cite{paulheimF12}. In particular, FeGeLOD extracts two types of features similar to ours: 
 paths of size 1 ($\xrightarrow[]{r} e$) starting at the entities of interest; paths of size 1 ($\xrightarrow[]{r} t$) where the original entity ($e$) is replaced with ontology classes ($t$) it instantiates. In the same vein, KGPTree~\cite{vandewiele19} extracts paths of the form \texttt{root} $\rightarrow$ \texttt{predicate} $\rightarrow$ \texttt{entity} $\rightarrow$ \texttt{predicate} \dots $\rightarrow$ \texttt{entity}, while allowing generalizations on both predicates and entities, whereas we offer generalization on entities only. 
 However, they only allow a generalization to a unique and broad type denoted with a wild card (*), while we allow a granular generalization following ontology class hierarchies.
 FeGeLOD and KGPTree extract only paths and path patterns, whereas one may want to extract other common structures such as subtrees. Mustard Python library offers such functionalities applying Graph Kernels on RDF graphs, plus additional facilities such as detecting hubs or low frequency patterns~\cite{vriesR13,vriesR15}. However, it does not allow generalization operations.
 
The contribution of our work is twofold: first, we show that knowledge graphs provide sufficiently diverse features to enable simple and explainable models to distinguish between drugs that are causative or not, for two types of ADRs commonly monitored; second, we evaluate manually that in this setting, predictive features constitute good candidates for investigating ADR mechanisms further.
The following sections present materials and methods, obtained results and a discussion about our methodological choices and results. 

\section*{Materials and methods}

\subsection*{Data sources}

\subsubsection*{PGxLOD}
PGxLOD is a linked open data (LOD) knowledge graph built for pharmacogenomics (PGx) and encoded in RDF~\cite{monninLHRTJNC19}.
It aggregates data mainly about drugs, genetic factors, phenotypes and their interactions from six data sources: PharmGKB, ClinVar, DrugBank, SIDER, DisGeNET and CTD; but also includes references to Gene Atlas, UniProt, GOA and KEGG. 
In particular, it includes pharmacogenomic relationships \textit{i.e.,} $n$-ary relations that represent how a genomic factor may impact a drug response phenotype. These relations are compiled from PharmGKB and the literature.
We used PGxLOD version 4 that encompasses 88,132,097 triples. 
Table \ref{tab:pgxlod_stats} presents its main statistics.
PGxLOD is available at \url{https://pgxlod.loria.fr}.

\begin{table}
        \centering
          \caption{\textbf{Types and numbers of entities available in the PGxLOD knowledge graph.} Pharmacogenomic relationships of PGxLOD are of two provenances: the PharmGKB expert database and the literature. }
	\begin{tabular}{lr}
		Concept                               & Number of instances  \\
		\hline
		\texttt{Drug}                           & 63,485  \\
		\texttt{GeneticFactor}                  & 494,982 \\
		\texttt{Phenotype}                      & 65,133  \\
		\texttt{PharmacogenomicRelationship}    & 50,435  \\
		\ \ \ \textit{from PharmGKB}            & 3,650   \\
		\ \ \ \textit{from the literature}      & 36,535  \\
	\end{tabular}
  
    \label{tab:pgxlod_stats}
\end{table}

In our study, PGxLOD enables to associate phenotypic and molecular features to drugs, by the exploration of their neighborhood in the graph.

\subsubsection*{Drug expert classifications and their preprocessing}

We experiment with two expert classifications of sets of drugs labeled as either causative or not for ADRs. The first concerns drugs causative for drug induced liver injury (DILI), and the second drugs causative for severe cutaneous adverse reactions (SCAR).

\paragraph{DILI classification} 
We built our DILI classification from DILIRank, a list of 1036 FDA-approved drugs classified by their risk of causing DILI~\cite{chen16}. DILIRank distinguishes between four classes listed in Table \ref{table:dili-classes}. This classification was obtained by \textit{(i)} the curation of information gathered from FDA-approved drug labels, setting an initial list of 287 drugs; \textit{(ii)} a semi-automatic approach that completes the list up to 1036 drugs, by combining information from hepatotoxicity studies and the literature. 
We sub-sampled from DILIRank \textbf{370 drugs (146 DILI$\oplus$, and 224 DILI$\ominus$)} that fulfill criteria required for our subsequent analysis: being either in the Most- or No- DILI concern classes, being associated with a SMILES (simplified molecular-input line-entry system) description, and being mapped to PGxLOD. The latter mapping was obtained with PubChemIDs, which are available both in DILIRank and PGxLOD (coming from PharmGKB, DrugBank, and/or KEGG).
Drugs satisfying these criteria and classified as Most-DILI concern constitutes the DILI$\oplus$ subset, and those classified as No-DILI concern constitutes the DILI$\ominus$. 


\begin{table}
    \centering
    \caption{\textbf{Classes and size of the original DILIRank expert classification.} Classes group drugs causative or not for drug-induced liver injury (DILI) on the basis of FDA-approved drug labels and a semi-automatic method. }
    \begin{tabular}{p{4cm}r}
        \hline
         Class &  \# drugs\\
         \hline
         Most DILI concern & 192 \\
         Ambiguous DILI concern & 254 \\
         Less DILI concern & 278 \\
         No DILI concern & 312 \\
         \hline
         \textbf{Total} & \textbf{1036} \\
         \hline
    \end{tabular}
    \label{table:dili-classes}
\end{table}

\paragraph{SCAR classification} 
Our SCAR classification relies on a manually built classification called ``Drug notoriety list", shared by members of the RegiSCAR project.
This list was originally assembled for the evaluation of the ALDEN algorithm, which assesses the chance for a drug to cause  Stevens–Johnson Syndrome
and Toxic Epidermal Necrolysis, a specific type of SCAR \cite{regiscar_notoriety_list, sassolas10}. This classification lists 874 drugs and assigns them to five classes representing various levels of association with SCAR. 
These classes are listed in Table  \ref{table:scar-classes}.
We sub-sampled from the RegiSCAR drug notoriety list \textbf{392 drugs (102 SCAR$\oplus$, and 290 SCAR$\ominus$)} drugs fulfilling two criteria: being mapped to PGxLOD and having a SMILES description. 
The mapping starts with 874 drug names, which is the only description available in the RegiSCAR list. Drug names are matched with lists of synonyms associated with drugs in PharmGKB and DrugBank.
Drugs satisfying these criteria and classified as Very probable, Probable or Possible constitute the SCAR$\oplus$ subset, and those classified as Unlikely or Very unlikely, the SCAR$\ominus$.

\begin{table}
    \centering
    \caption{\textbf{Classes and size of the original RegiSCAR drug notoriety list.} Classes group drugs causative or not for severe cutaneous adverse reactions (SCAR).}
    \begin{tabular}{p{4cm}r}
        \hline
         Class &  \# drugs\\
         \hline
         Very probable (3) & 18 \\
         Probable (2) & 19 \\
         Possible (1) & 94 \\
         Unlikely (0) & 697 \\
         Very unlikely (-1) & 46 \\
         \hline
         \textbf{Total} & \textbf{874} \\
         \hline
    \end{tabular}
    \label{table:scar-classes}
\end{table}

\subsection*{Methods}

\begin{figure}[htb!]
    \centering
    \includegraphics[width=10cm]{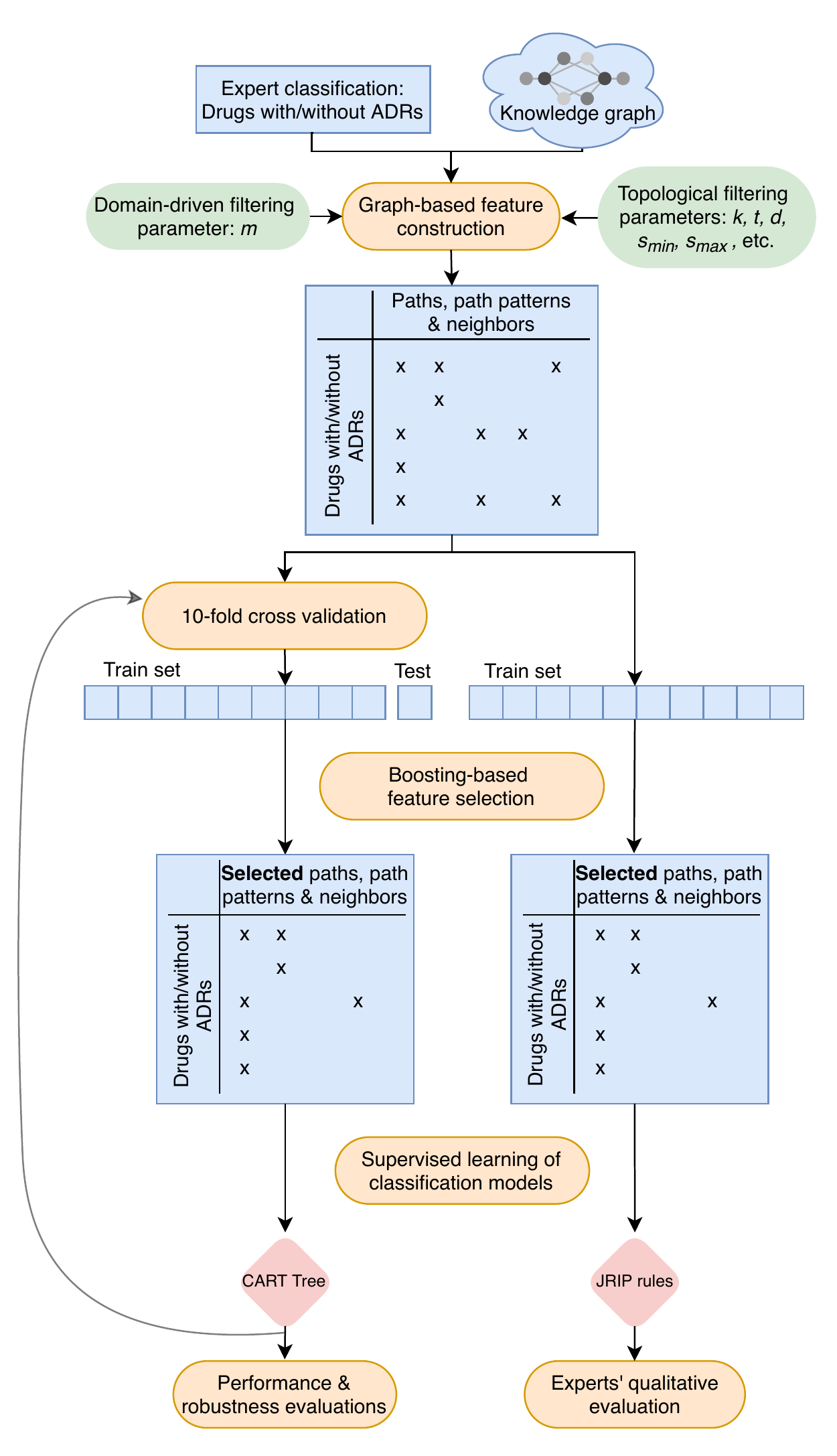}
    \caption{\textbf{Method overview.} Main steps are in orange, data in blue, parameters in green and models in pink.}
    \label{fig:global-approach}
\end{figure}

An overview of the steps of the proposed method  is provided in Figure~\ref{fig:global-approach}.

\subsubsection*{Graph-based feature construction}

\paragraph{Graph canonicalization}
In knowledge graphs, several nodes may co-exist, while representing the same entity.
For example, a unique drug in PGxLOD can be represented by two nodes: one from PharmGKB and another from DrugBank.
Each of these two nodes have their own connections to other nodes in the knowledge graph.
Therefore, the union of their edges constitutes all the available knowledge about the drug they both represent.
As they represent the same drug, they are connected through an \texttt{owl:sameAs} edge.
In order to avoid traversing such edges, we \textit{canonicalize} the knowledge graph, that is to say, nodes linked by \texttt{owl:sameAs} edges are grouped into a unique node as a pre-processing step, before building graph-based features.
Such a procedure corresponds, in graph theory, to the contraction of \texttt{owl:sameAs} edges.

\paragraph{Paths, path patterns, and neighbors as drug features}

From an arbitrary set of drugs $D$ (such as DILI$\oplus$ $\cup$ DILI$\ominus$) and a canonicalized knowledge graph (such as PGxLOD), a set of graph-based features is built on $D$. 
We distinguish three kinds of features: paths, path patterns and neighbor nodes. 
First, we build \textbf{paths} rooted by drugs from $D$. 
The neighborhood of each drug $d \in D$ is explored in the graph with a max distance of $k$, generating sequences of predicates and nodes of max length $k$.
Accordingly, if a path $d \xrightarrow{p_1} n_1 \xrightarrow{p_2} n_2$ is found in the knowledge graph (with $k=2$), the drug $d$ is associated with the feature $\xrightarrow{p_1} n_1 \xrightarrow{p_2} n_2$ in the output matrix.

Second, we build \textbf{path patterns} that generalize paths by considering ontology classes instantiated by nodes. The aim of path patterns is to offer more general descriptions, which have more chances to be shared by several entities. 
For example, if $n_1$ instantiates $C_1$ and $n_2$ instantiates $C_2$, we add the following path patterns: 
$\xrightarrow{p_1} C_1 \xrightarrow{p_2} n_2$, $\xrightarrow{p_1} n_1 \xrightarrow{p_2} C_2$, $\xrightarrow{p_1} C_1 \xrightarrow{p_2} C_2$, $\xrightarrow{p_1} \top \xrightarrow{p_2} n_2$, $\xrightarrow{p_1} n_1 \xrightarrow{p_2} \top$, $\xrightarrow{p_1} C_1 \xrightarrow{p_2} \top$, $\xrightarrow{p_1} \top \xrightarrow{p_2} C_2$, and $\xrightarrow{p_1} \top \xrightarrow{p_2} \top$.
It is noteworthy that, to allow a high level of generalization, we always consider the top-level class $\top$ in the generalization procedure.
To leverage hierarchies organizing ontology classes, a node $n$ is only generalized by ontology classes at a distance of at most $t$ from itself, following instantiation and subsumption edges.

Third, we list \textbf{neighboring nodes}, \ie any node that can be reached from $d \in D$ within a distance of max $k$.
We do not keep track of the distance between $d$ and its neighbors: if a node $n$ can be reached from a drug $d_1$ after 2 hops and from a drug $d_2$ after 3 hops, these two drugs will be associated with the neighbor $n$ in the output matrix.
Interestingly, a neighbor $n$ can be represented by the general path pattern $(\xrightarrow{*}*)^{\{0,k-1\}}\xrightarrow{*}n$.

Features can potentially be noisy and numerous, leading to a combinatorial explosion of their number.
That is why, we add several constraints to only keep the most interesting features.
First, we only keep the most specific paths and path patterns, considering that a node is more specific than the classes it instantiates and a class is more specific than its superclasses.
A path or path pattern $P_1$ is more specific than another path or path pattern $P_2$  if each node/class in $P_1$ is more specific than the node/class at the same position in $P_2$.
Thus, if $\xrightarrow{p_1} n_1 \xrightarrow{p_2} n_2$ is associated with the exact same drugs as $\xrightarrow{p_1} C_1 \xrightarrow{p_2} C_2$, then the path pattern is removed and only the path constitutes a feature.
Similarly, if $\xrightarrow{p_1} C_1 \xrightarrow{p_2} C_2$ is associated with the exact same drugs as $\xrightarrow{p_1} \top \xrightarrow{p_2} \top$, then we only keep the first one.
Second, the exploration is stopped at nodes whose degree is greater than a parameter $deg$.
Such nodes are usually called \textit{hubs} \cite{vriesR15}, and 
expanding a path ending at a hub may generate a large number of new paths associated with the same drugs. 
Third, we consider minimal and maximal \textit{supports} of features (denoted $s_{\text{min}}$ and $s_{\text{max}}$) as additional filters. The support of a feature consists in the number of drugs associated with a feature, \ie neighbors of a same entity or rooting a path or a path pattern. 
Only features associated with more than $s_{\text{min}}$ drugs and less than $s_{\text{max}}$ drugs are output in the final matrix.
Fourth, two blacklists avoid traversing noisy or unwanted edges.
Edges are not traversed if their predicate is blacklisted (in $b_{\text{predicates}}$) or if the adjacent node instantiates (directly or indirectly) a blacklisted class (in $b_{\text{exp-types}}$).
For example, we blacklist predicates of PROV-O that are used to describe provenance metadata.
By blacklisting \texttt{rdf:type} in $b_{\text{predicates}}$, we make sure the exploration is performed through entities and not classes of PGxLOD. 
Relations between classes are only considered when generalizing paths.
Aside from noisy features and combinatorial explosion, we use these blacklists to prevent considering features that ``obviously'' carry the signal we are trying to predict.
For example, we blacklist all ``side-effects'' links from SIDER, which may directly link drugs in $D$ with the side effect we aim at predicting.
We also blacklist all classes from MeSH related to SCAR or DILI to avoid taking into account nodes instantiating them.
A third blacklist ($b_{\text{gen-types}}$) avoids generalizing nodes in paths by blacklisted classes.
This is particularly useful to withdraw general classes (\eg \texttt{Drug}) that increase the number of generated path patterns while not adding useful information.

Besides previous topological constraints, we also perform a domain-driven filtering configured by parameter $m$.
Indeed, in our objective of explaining ADRs, experts highlighted that features that may be explanatory to them mention pathways, genes, GO, and MeSH terms.
For this reason, we propose three post-processing atomic filters, only keeping neighbors, paths or path patterns containing at least a pathway ($m=\mtt{p}$), a gene or a GO class ($m=\mtt{g}$), or a MeSH class ($m=\mtt{m}$).
Such atomic filters can be combined to form disjunctive filters.
For example, the $m=\mtt{pg}$ filter keeps neighbors, paths or path patterns containing at least a pathway or a gene or a GO class.

Table \ref{table:graph-based-parameters} summarizes the parameters that limit the number of features in the output matrix. Additional details about our method of feature construction are provided in \cite{monnin:hal-02913224}. 

\begin{table}
	\caption{\textbf{Parameters used to limit the number of features.}
	Each parameter is associated with a domain for its value.
	}
	\begin{center}
		\begin{tabular}{lp{2.2cm}p{7.5cm}}
			\hline
			Parameter & Domain & Description \\
			\hline
			$k$ & $\mathbb{N}^+$ & Maximum length of paths and path patterns\\
			$t$ & $\mathbb{N}\cup\left\{-1\right\}$ & Maximum depth of generalization\\
			$deg$ & $\mathbb{N}\cup\left\{-1\right\}$ & Maximum node degree to allow expansion\\ 
			$s_{\text{min}}$ & $\mathbb{N}$ & Minimum support for features\\ 
			$s_{\text{max}}$ & $\mathbb{N}$ & Maximum support for features\\ 
			\small{$undirected$} & $\mathbb{B}$ & Consider the graph undirected or directed\\
			$b_{\text{predicates}}$ & List of URIs 
			& Blacklist of predicates not to traverse \\
			$b_{\text{exp-types}}$ & List of URIs 
			& Blacklist of types of entities not to traverse \\
			$b_{\text{gen-types}}$ & List of URIs 
			& Blacklist of types not to traverse in generalization \\
			$m$ & $\left\{\mtt{no-check}\right.$, $\mtt{p}$, $\mtt{g}$, $\mtt{m},$ $\mtt{pg}$, $\left.\mtt{pgm}\right\}$ & Domain-dependent filter for features with at least a pathway (\mtt{p}), a gene (\mtt{g}) or a MeSH class (\mtt{m})\\
			\hline
		\end{tabular}
	\end{center}
	\label{table:graph-based-parameters}
\end{table}

\subsubsection*{Cross-validation strategy}

Once drug features are extracted from the knowledge graph, they are given to a machine learning algorithm that learns a model, which mimics the expert classification. To quantitatively evaluate our approach, we adopt a 10-fold cross-validation strategy, meaning that we repeated the following steps of our learning pipeline (\textit{i.e.,} feature selection, and training) 10 times, holding out each time one tenth of our labeled data for testing. The split in 10 folds is performed once, randomly. 

\subsubsection*{Feature selection with boosting}

The concept of boosting relies on the sequential learning of classifiers that successively focus on getting correct the examples that were wrongly classified by previous classifiers, with a weight system: correctly classified examples loose weight, whereas falsely classified ones gain weight \cite{kearns1988}. The final classification is built by a weighting of the results obtained from the various classifiers. 
Ensembles of decision trees can be used in a boosting strategy for the estimation of feature importance in order to select the most important features for a subsequent learning model \cite{adaboost_for_feature_selection}. This approach is frequently named ``wrapper-based feature selection''. 
We adopt this approach, using the AdaBoost algorithm, with up to 10 decisions trees learned from the whole train set \cite{adaboost}. In case of perfect fit, the learning procedure is stopped early. For the Decision Tree algorithm we used CART from scikit-learn, Gini impurity, all features considered at each split and a $minimal\ number\ of\ examples\ per\ leaf = 5$. Every train example is used, but classes are artificially balanced: examples are associated with a different weight depending on whether they belong to the $\oplus$ or $\ominus$ class. Features that appear in at least one of the decision trees are selected for the subsequent step of our learning pipeline. 

\subsubsection*{Performance and robustness evaluation}

For the evaluation of our capacity of distinguishing between drugs associated or not with an ADR, we train a last decision tree, using selected features only. Note that because the selection step is repeated at each iteration of the cross validation strategy, selected features may vary from one iteration to another. Algorithm and parameters are the same as those of the selection step, \ie CART from scikit-learn with Gini, all features considered at each split, $minimal\ number\ of\ examples\ per\ leaf = 5$, and weighted instances for class balancing.

\paragraph{Performance results} are reported in term of Precision, Recall, 
F1-score (reference class $\oplus$), accuracy and AUC-ROC. Metrics are averaged over the 10 iterations of the cross validation. 

\paragraph{A robustness evaluation} is also performed to assess the impact of the train set (\ie expert classifications) on the final classifier. To this aim, first, we reproduced the experiment, but with a shuffled class assignment in the train set. This leads to a train, denoted by $\cdot^{shuffled}$ (where $\cdot$ is either $DILI$ or $SCAR$) where drugs are associated randomly to either the class $\oplus$ or $\ominus$.
This first sanity evaluation mainly checks the presence of a nonrandom signal in the train. Second, we replace the set of drugs from $\ominus$ by a set of drugs randomly picked out of our knowledge graph. We repeat this random draw five times for each expert classification resulting in 10 train sets denoted $\cdot^{random \ominus_{i}}$ (where $\cdot$ is either $DILI$ or $SCAR$ and $i$ is an index taking values from 1 to 5). In each case the draw is made from nodes of the knowledge graph that instantiate (directly or indirectly) \texttt{pgxo:Drug}; are identified with a URI from PharmGKB or DrugBank namespaces; are linked by a \texttt{x-pubchem} predicate to a PubChem URI (in order to have a SMILES associated with the drug); and are not drugs of the original DILI or SCAR expert classifications. 
In the case of DILI, this is a draw of 224 nodes out of 5893 in the canonical graph. In the case of SCAR, this is a draw of 290 nodes out of 5921.
This second evaluation checks the impact of the selection of negative examples on classifier performances. 
In addition, we count the number of features that are present in 5, 4, 3 or 2 of the five $\cdot^{random \ominus_{i}}$ experiments.  

\subsubsection*{Qualitative evaluation by human experts}

To go beyond performance evaluation, we produce a classification model of rules using the RIPPER algorithm, and its Java implementation named JRip \cite{cohen1995}. 
JRip has the advantage over decision trees to provide classification rules that are more concise and by consequence easier to interpret for humans. JRip actually implements a propositional rule learner,  produces relatively non redundant rules in comparison to rules that could be learned following branches of a decision tree. However, JRip and CART decision tree usually perform very similarly, since they implement similar pruning strategies and stopping criteria. To evaluate this assessment, we performed a 10-fold cross validation of the JRip approach and compared performances with CART. 
JRip generates rules of the form: 
$$ \displaystyle \left(\bigwedge_{\forall a \in A} a \right) \wedge \left( \bigwedge_{\forall b \in B} \neg b \right) \Rightarrow c $$ where $A$ is a set of attribute-value pairs affirmed, $B$ is a set of attribute-value pairs negated, and $c$ is the minority class of the classification problem. Accordingly, $c=\oplus$ in our study. Note that $A$ or $B$ can be an empty set, but not both at the same time. In other words, JRip rules consist in the conjunction of affirmed and negated features.

As for CART Trees, we start with an initial step of feature selection with AdaBoost, with the same parameters but this time considering all examples of the train set.
Following, JRip rules are built considering also all examples of the train set, and features that appear in at least one of the trees built by AdaBoost. 
To be consistent, with the CART Tree setting, we set to 5 the \textit{minimal number of instances per rule}, which can be compared to the \textit{minimal number of examples per leaf}. 

 JRip rules are post-processed to facilitate their interpretation by our experts. 
 First, among features that are path patterns, we discard those only involving generic classes such as \texttt{Resource} or \texttt{Drug}.
 Indeed, such path patterns turned out to be impossible to interpret.
 Second, features are translated into a readable format, by resolving ids with associated labels (\eg \texttt{drugbank:BE0003543} is resolved as ``Cytochrome P450 1A1") and by interpreting and rewriting paths and path patterns in an understandable form (\eg $\xrightarrow{\texttt{drugbank\_vocabulary:enzyme}}$ $\texttt{\small genatlas\_vocabulary:Resource}$ is turned in $\xrightarrow{\text{interactsWith}}$ $\text{Enzyme}$). 
 Because of the limited number of features in rules, this translation is made manually, on the basis of descriptions of predicates, classes, and entities found in their original knowledge graph or database.

 We asked three experts in pharmacy and pharmacology to review independently each attribute (\ie feature) of the rules, to evaluate if they may be explanatory for ADRs. Each expert has to answer a voluntarily simple three-way question: ``according to your own knowledge or the state of the art, do you think that the feature is explanatory for DILI?'' (SCAR, respectively). Possible answers are ``yes'', ``maybe (possible, but not obvious)'' and ``no (probably not explanatory)''. We allow expert to check the literature or any state-of-the-art resource, but up to 15 minutes, since we consider that more time causes to fall in the ``no (probably not explanatory)'' option. To guide their decision on each feature, experts are provided with two Web links: one pointing to the list of drugs from the $\oplus$ train set that supports the feature; one to the page of the main entity mentioned in the feature (\ie the neighbor node, or the final node of the path or path pattern) in an expert database: DrugBank, ChEBI, KEGG, QuickGO, or MeSH browser, depending on the namespace of the node. 
 After expert reviews, answers were normalized, under their supervision, to guarantee all experts interpret the negation of features the same way. 
 For each feature we check if at least one, two or three of the experts think it is or may be explanatory,  if the three think it is explanatory and if the three think it is not. 
In addition, we compute Cohen's kappa coefficient to evaluate the average agreement between experts with two different settings: considering the three different answers as distinct, or considering answers ``yes'' and ``maybe'' as a unique positive answer. 

\section*{Results}

\subsubsection*{Graph-based feature construction}

We experimented graph exploration with combinations of the following parameters values $k \in \{1,2,3,4\}$, $t \in \{1,2,3\}$, $deg = 500$, $undirected = \mtt{false}$, $s_\text{min} = 5$, $s_\text{max} = +\infty$ and $m \in \{\mtt{p}, \mtt{g}, \mtt{m}, \mtt{pg}, \mtt{pgm}\}$.
$k = 4$ was only tested with $t = 1$ because of memory issues caused by the high number of generated features with greater values of $t$.
However, we report only the best results, which were obtained with $k = 3$, $t = 3$, $m = \mtt{pgm}$ for DILI and $m = \mtt{pg}$ for SCAR.
These values of $m$ enable to conserve only features that includes an entity that is either a pathway, a gene, or a GO term (for $\mtt{pg}$); or a pathway, a gene, a GO, or MeSH term (for $\mtt{pgm}$).

The construction of graph-based features is obviously limited by the amount of available memory.
We enforce $s_\text{min}$ to be set  
to allow the construction of paths and path patterns, in order to avoid combinatorial explosion. Accordingly their number is reported only once we reduced the number of all possible combinations, which we were not able to compute.
We used a server with a Xeon E5-2680 v4@2.40GHz CPU, 28 cores/56 threads
and 768GB of memory. 
As an illustration, we obtained the features associated with the DILI expert classification under $k = 3$, $t = 3$ in approximately 1 hour using 62 GB of RAM, and under $k= 4$, $t= 1$ in 4 days using 380 GB of RAM. 

To provide with an idea of the size of the considered neighborhood with regards to all reachable nodes, Table~\ref{tab:graph-feature-results} reports statistics about numbers of neighbors, paths and path patterns reachable with different level of filtering. 
In particular, we report sizes of the full neighborhood of drugs, and of 3 levels of filtering. 
The first level of filtering consists in prohibiting the expansion of the neighborhood through nodes with a degree higher than 500 ($deg=500$).
In the full neighborhood and first level of filtering, $k$ and $t$ are not constrained since no path or path pattern is computed, but we report max $k$ and $t$ reached in the neighborhood. We note that $k$ is surprisingly lower in the larger neighborhood, \ie 19 and 20 vs. 23 and 23 with the first level of filtering. This can be explained by the fact that with ``hubs'' (\textit{nodes with $deg > 500$}), the full neighborhood can be reached through smaller paths. We also observe that because of this first filtering, certain nodes are not accessible anymore (when every possible path to them pass through a hub), which results in a smaller number of neighbors.
The second level of filtering comes on top of the first, and constrains neighbors, paths, and path patterns to have a minimal support set of 5 ($s_\text{min} = 5$), a max length of 3 ($k=3$), and a max depth 3 for generalization of paths into patterns ($t=3$).
The filtering level 3 comes on top of the second, and constrains neighbors, paths, and path patterns to contain a pathway, a gene, a GO term, or a MeSH term for DILI ($m = \texttt{pgm}$), or to contain a pathway, a gene, or a GO term for SCAR ($m = \texttt{pg}$).
The filtering level 3 is the one used in the following experiments,  because it is computable in our setting, while providing the best performances in our set of experiments.

\begin{table}[]
    \centering
    \caption{\textbf{Numbers of drug features extractable from the knowledge graph, with different levels of filtering.} 
    The first line corresponds to the full neighborhood of drugs from DILI and SCAR expert classifications. 
    $deg = -1$ means that all nodes are considered, regardless of their degree, whereas $deg=500$ in Filtering level 1 means that nodes with a degree $> deg$ are filtered out.
    In the two first lines (No filtering and Filtering level 1), $k$ and $t$ are unconstrained, so reported values are maximum $k$ and $t$ observed in the graph.
    Paths and paths pattern are computed only when $deg$ and $s_\text{min}$ (minimum support) are set, to avoid combinatorial explosion. 
    Filtering level 2 and 3 share the following additional parameters: $undirected=\mtt{false}$, $s_\text{max}=+\infty$.
    In Filtering level 3, $m$ is set for additional filtering. Distinct values for $m$ chosen respectively for DILI and SCAR are those associated with the best performances, \eg $m_{DILI}=\mtt{pgm}$ and $m_{SCAR}=\mtt{pg}$. }
    \begin{tabular}{p{0.38\textwidth}p{0.22\textwidth}cc}
        \hline
        & & $DILI$ & $SCAR$ \\
        \hline
        & Neighbors & 5,488,531 & 5,488,510 \\
        No filtering & $k$ & 19 & 20 \\
        \ \ ($deg$=-1)& $t$ & 21 & 21 \\
        \hline
        & Neighbors & 2,419,957 & 2,419,920 \\
        Filtering level 1 & $k$ & 23 & 23 \\
        \ \ ($deg$=500) & $t$ & 21 & 21 \\
        \hline
        Filtering level 2  & Neighbors & 175,652 & 179,694 \\
        \ \ \small{($deg$=500, $s_\text{min}$=5, $k$=3, $t=$3)} & Paths \& path patterns & 20,145,635 & 29,011,996\\ 
        
        \hline
        Filtering level 3  & & & \\
        \ \ \small{($deg$=\textbf{500}, $s_\text{min}$=\textbf{5}, $k$=\textbf{3}, $t=$\textbf{3},} & Neighbors & 4,069 & 1,594 \\
                \ \ \small{$m_{DILI}$=\texttt{\textbf{pgm}} and $m_{SCAR}$=\texttt{\textbf{pg}})}& Paths \& path patterns & 102,674 & 86,753 \\
        \hline

    \end{tabular}
    \label{tab:graph-feature-results}
\end{table}

\subsection*{Quantitative and robustness evaluation}

Performances of our CART decision trees to distinguish between drug associated or not with ADRs are reported Table \ref{table:quantitatve-evaluation}. With both types of ADRs, we obtained accuracy and AUC higher than 0.70, illustrating the fact that learned classifiers reproduced a large part of expert classifications, on the basis of features of the knowledge graph. 

Robustness of classifiers is illustrated by the results provided in Table \ref{table:robustness-evaluation}. $DILI^{shuffled}$ and  $SCAR^{shuffled}$ are associated with AUC of 0.49 and 0.51, respectively. This illustrates that a random assignment of class labels in the train set, leads to a classifier that randomly assign labels to test examples. This is expected, but illustrates that expert classifications encompass a signal that our classifiers learn and reproduce, to some extent. 

Classifiers trained with a random pick of negative examples ($\cdot^{random \ominus_{i}}$) instead of negatives picked by experts are significantly better for the three performance metrics (T-Test, $t > 36$, $p <1.7 10^{-6}$). This reveals it is harder for our classifier to discriminate between positives and negatives of expert classifications, than it is between positives and randomly picked drugs. This lets us assume that negatives from expert classifications are somehow similar to positives (they may share some properties) and harder to distinguish for the classifier, even if not associated with the studied ADR. 
When comparing features used in the five random pick of $DILI^{random \ominus_{i}}$, we observed that, out of a mean of 122 features (sd=8), $6$, $12$, $27$ and $99$ were common to respectively 5, 4, 3 and 2 picks. With $SCAR^{random \ominus_{i}}$, out of 108 features (sd=8), $2$, $6$, $23$ and $81$  were common to respectively 5, 4, 3 and 2 picks.

\begin{table}
    \centering
    \caption{\textbf{Quantitative evaluation of our classifiers of drugs associated with ADRs or not (DILI or SCAR).}}
    \begin{tabular}{llccccccc}
        \hline
         Algorithm &  Data set & Precision & Recall & Accuracy & AUC & F1-score\\

         \hline
         \multirow{2}*{CART} & $DILI$ & 0.68 
         & 0.67 
         & 0.74 & 0.73 & 0.67 \\
         & $SCAR$ & 0.64 
         & 0.68 
         & 0.81 & 0.77 & 0.65\\
         \hline
         \multirow{2}*{JRip} & $DILI$ & 0.82 
         & 0.71 
         & 0.72 & 0.74 & 	0.75 \\
         & $SCAR$ & 0.88 
         & 0.70	
         & 0.71 & 0.74 & 0.77\\
         \hline
    \end{tabular}
    \label{table:quantitatve-evaluation}
\end{table}

\begin{table}
    \centering
    \caption{\textbf{Robustness evaluation of our classifiers.} $\cdot^{shuffled}$ corresponds to an experiment where class labels (\ie $\oplus$ or $\ominus$) are randomly affected to drugs. $\cdot^{random \ominus_{i}}$ correspond to experiments where negative examples (\ie $\ominus$) are replaced by drugs randomly picked in the knowledge graph. Indices $i$ from 1 to 5 refer to 5 different draws.}
    \begin{tabular}{lccc}
        \hline
         Data set & Accuracy & AUC & F1-score $\oplus$\\
         \hline
         $DILI^{shuffled}$  & 0.52 & 0.49 & 0.36 \\
         $DILI^{random \ominus_1}$ & 0.92 & 0.91 & 0.89 \\
         $DILI^{random \ominus_2}$ & 0.92 & 0.91 & 0.89 \\
         $DILI^{random \ominus_3}$ & 0.93 & 0.92 & 0.91 \\
         $DILI^{random \ominus_4}$ & 0.93 & 0.92 & 0.90 \\
         $DILI^{random \ominus_5}$ & 0.92 & 0.91 & 0.90 \\
         \hline
         $SCAR^{shuffled}$ & 0.63 & 0.51 & 0.26\\
         $SCAR^{random \ominus_1}$ & 0.93 & 0.89 & 0.86 \\
         $SCAR^{random \ominus_2}$ & 0.94 & 0.90 & 0.88 \\
         $SCAR^{random \ominus_3}$ & 0.93 & 0.90 & 0.86 \\
         $SCAR^{random \ominus_4}$ & 0.92 & 0.89 & 0.85 \\
         $SCAR^{random \ominus_5}$ & 0.93 & 0.89 & 0.86 \\
         \hline
    \end{tabular}
    \label{table:robustness-evaluation}
\end{table}

\subsection*{Expert evaluation}

JRip produced 6 and 5 rules for DILI and SCAR, respectively. 
The translation of these rules is available in Supplementary Tables S\ref{tab:s1} and S\ref{tab:s2} (see Additional information section). 
After removing, uninformative features, we obtained 11 and 13 distinct features, respectively. Those are provided in Supplementary Tables S\ref{tab:s3} and S\ref{tab:s4} (see  Additional information section). These features are those reviewed by our three experts in pharmacology (CB, CNC, and NP). Quantitative performances of the JRip algorithm are presented in Table \ref{table:quantitatve-evaluation} for comparison with CART. On both datasets (DILI and SCAR) differences in performances (Precision, Recall, Accuracy, F-measure) are not statistically significants (T-Test, $p<0.05$).

The ratio (and number) of features for which experts reached an agreement, or for which 
at least one, two or three of the experts think they are or may be explanatory (answers ``yes'' or ``maybe'') are provided in Table \ref{tab:agreement-like}. We observe that  no feature generated by JRip is considered as unexplanatory by all three experts. In other words, every feature is thought as possibly explanatory by at least one expert. 
The ratio of features having a full agreement between experts on the possibility of being explanatory is reduced compared to those having a partial agreement but stays relatively high (0.73 features) for DILI and moderate (0.38 features) for SCAR. 
Full agreement for features being explanatory (all three experts answer ``yes'') remains minor, but exists. 
Kappa's Cohen agreement score is $\kappa_{n=3} = 0.26$ when considering answers ``yes'', ``maybe'', and ``no'' independently, but reaches $\kappa_{n=2} = 0.70$ when the problem is reduced to two classes by aggregating ``yes'' and ``maybe'' answers. 
Note that Table \ref{tab:agreement-like} reports in its fifth column the ratio of features that reach full agreement for our three experts when ``yes'' and ``maybe'' answers are aggregated. Additional file 1, available upon request, contains results of the manual evaluation of the features.

\subsection*{Examples of features and elements of interpretation}

Three features reach an agreement for being explanatory (\ie three answers ``yes'' per feature). Those can be interpretated as elements that are well known for being explanatory, or at least associated, with DILI or SCAR mechanisms. 

As a first illustration, 
$\xrightarrow{\textbf{interactsWith}} \textbf{Enzyme} \xrightarrow{\textbf{cellularComponent}} \textbf{Endoplasmic}$ $\textbf{reticulum }$ reached an agreement for DILI. This is explained by the fact that endoplasmic reticulum is known, in particular in liver tissues, to host primarily cytochrome P450 enzymes, well known for being involved in drug metabolism \cite{neve2010}. 

As a second illustration, $\textbf{Cytochrome P450 2B6}$ reached an agreement for SCAR, whereas genomic variations in the gene coding for this enzyme are known for being associated with SCAR~\cite{ciccacci2013}. One might consider fairly that this feature does not bring new explanatory elements, although it can be considered a minimum that our method highlights well established explanatory elements. 

All other features did not reach an agreement, or reach one for ``maybe''. Each of those is interesting to explore for further interpretation, but for the sack of briefness, we will only discuss two of them here. First, 
$\neg ( \xrightarrow{\textbf{involvedIn}} \textbf{Pathway} \xrightarrow{\textbf{associatedWith}} \textbf{Disease} \xrightarrow{\textbf{interactsWith}} \textbf{Calcium signaling pathway})$ reaches an agreement for potentially being  explanatory (\ie three answers ``maybe'') for DILI. This path pattern is relatively complex to interpret since it is long ($k=3$) and negated. Experts searched for literature reporting associations between DILI and Calcium signaling pathway. They found that a relatively old literature (old is seen as lacking confirmation by some experts) were reporting such an association \cite{jones1998}. A more recent bioinformatics article by Chen \textit{et al.}, also reported such association, but with a finer grain of information, since they report an association with hepatomegaly (a secondary example of DILI), and a negative association with hepathitis (a primary example of DILI) \cite{chen10}. Accordingly, negative results from this study are consistent with our finding of this latter negated feature. 
We note that Chen \textit{et al.} study is computational, as is ours, and that we may also be impacted by similar bias. 
Second, $\xrightarrow{\textbf{interactsWith}} \textbf{Enzyme} \xrightarrow{\textbf{biologicalProcess}}  \textbf{Electron transport}$ for SCAR, obtained very diverse opinions, with one ``no'', one ``maybe'' and one ``yes''. Even if this disagreement could be perceived as inconclusive, it may also point to a promising candidate for explanation. In this very one case, Electron transport is known for being perturbed in mitochondrion in many types of ADRs, including hepatotoxicity \cite{vuda2016}. However, we did not find any study reporting a link with skin toxicities.

 \begin{table}
    \centering
    \caption{\textbf{Ratio of features that either reach a full agreement for being unexplanatory, explanatory or are considered as possibly explanatory to various extents}. Absolute numbers are reported in parentheses. Ratio of unexplanatory features are in the left column, whereas explanatory features are in the right columns. The three middle columns count numbers of features that are, or may be, explanatory according to at least one, two or three experts. Numbers of considered features are 11 and 13 for DILI and SCAR respectively.}
    \begin{tabular}{p{0.08\textwidth}|c|ccc|c}
          & \multicolumn{5}{c}{\small{features }} \\
          & \small{with agreement}& \multicolumn{3}{c|}{\small{\textbf{possibly explanatory} for }} & \small{with agreement} \\ 
          \small{Data}&\small{on }& $\geqslant$ 1  & $\geqslant$ 2  & $\geqslant$ 3 & \small{on } \\
          
        \small{set}&\small{\textbf{unexplanatory } }&\multicolumn{3}{c|}{\small{experts}} & \small{\textbf{explanatory} } \\ 

         \hline
         $DILI$ & 0 & 1 (11) & 0.90 (10) & 0.73 (8) & 0.18 (2) \\
         $SCAR$ & 0 & 1 (13) & 0.77 (10) & 0.38 (5) & 0.08 (1)\\
         \hline
    \end{tabular}
    \label{tab:agreement-like}
\end{table}

\section*{Discussion}

In our work, simple, but explainable, classifiers (CART Decision Trees and JRip) were preferred to more advanced machine learning methods. Even if we are convinced that methods based on deep neural networks, such as graph embedding with Graph Convolutional Networks (GCN), should lead to better performances than those obtained \cite{kipfW16,schlichtkrullKB18}. However, acquiring explanatory elements about decisions made by such models necessitates a additional step of neural network analysis, such as saliency maps~\cite{mundhenk2019}, which provides information such as the layer or neurones activated by some instances. We consider this information of high interest for data scientists, but such methods require high level interpretation before being understandable by typical domain experts, unfamiliar with neural networks \cite{ying2019}. Consequently, such direction did not seem mature enough to reach our objectives. For instance, heatmaps that are to some extent explanatory for image classification, are still hard to transpose to knowledge graphs \cite{montavon2018}. However, it would be of interest to evaluate performances of a GCN on the classification task to measure the gap caused by our choice of simple classifiers. We also hope that our work will motivate studies on explainable subsymbolic approaches.

Our approach is reproducible for other applications. For instance, the same rational could be applied for the investigation of the mechanism of drug beneficial effects. This would necessitate to change our expert classifications for lists of drugs with a same indication ($\oplus$) and drugs without effect for this indication ($\ominus$), which could obtained from Sider or DailyMed. 

An objective of our work is to illustrate various advantages of mining knowledge graphs, and particularly semantic web ones. First, they provide human-readable features, that may subsequently be interpreted by experts. Second, predictive features may come from various connected data sets and jointly used in a single rule, which would have not been found if data sets were considered isolated. In addition, using semantic web standards eases the addition to our graph of novel data, following other \texttt{owl:sameAs} links. Third, semantic web knowledge graphs are associated with a formal semantics we benefit from at two steps: at the initial canonicalization, and at the generalization of path patterns. In this regards, one may ask if we could benefit from additional reasoning mechanisms, such as generalization over predicates. In our very specific case, predicates are not associated with any hierarchy, so it would not have changed our results, but from a general point of view, path patterns would benefit from this mechanism. Similarly, we could think of a canonicalization, not only with \texttt{owl:sameAs} links, but also following properties carrying similar semantics (\eg \texttt{skos:exactMatch}) or by applying matching approaches such as PARIS~\cite{suchanekAS11}.

When testing with values between 1 and 3, we observed that larger $t$ and $k$ are associated with better performances.
To achieve this, we adopted a rational approach for scaling the mining of RDF knowledge graph, which is presented in \cite{monnin:hal-02913224}. This approach reaches its limits with $k>3$ and $t>3$, but we think that additional optimization in the graph mining is still possible and would enable going further.

We used only binary features as they are easy to consider as explanations. Other strategies 
(\eg counting, relative counting~\cite{ristoskiP14pkdd}) could also have been considered while maybe hindering the descriptive power of such features. To maximize the descriptive power of candidate features and avoid redundancy, one could use specific metrics  (\eg approaches relying on hierarchies~\cite{ristoskiP14} and/or extent of classes~\cite{dAmatoSF08} of ontologies). 
Such metrics could also be considered within the decision tree algorithm, to propose to the algorithm additional features (more or less aggregated according to generalization) that may be associated with best split with regards to the Gini index (or others). This would lead to the consideration of formal knowledge directly in the mining algorithm~\cite{ristoskiP16}. Also, we proposed, with our parameter $m$ an hardcoded way of selecting features of interest. However one may think of an interactive selection by user, following the possibilities  offered by an ontology. 

A usual difficulty in human annotation or human evaluation is the normalization of expert answers. Despite a one-hour training about the task, the interpretation of negated features has been heterogeneous among experts. One pitfall was to think that if the affirmation of a feature is true, its negation is wrong. This is misleading because it is possible that a feature is explanatory for some examples and that its negation is also explanatory for other examples or in another context. To ensure normalization of negated features, we considered a feature as explanatory if its affirmation or its negation is explanatory to the expert. This change has been considered in a normalisation batch of reviews done in cooperation with experts.

Our review by human experts evaluates how many features highlighted by our approach are relevant (similarly to what Precision measures), but does not evaluate how many relevant features we may miss (similarly to what Recall measures). It would be of interest to ask experts what are features such as pathways, drug targets, cellular functions that are known to be associated with DILI and SCAR ADRs to enable a final comparison. However, establishing an exhaustive list from the state of the art would be complex and time consuming for experts. Text mining approaches could be of interest to guide them in this matter. 

\section*{Conclusion}

We illustrate in this work that life science knowledge graphs provide sufficiently diverse features to enable simple and explainable models to distinguish between drugs that are causative, or not, for two severe ADRs. These features take the form of paths, path patterns or simple neighboring nodes from the graph, which have the advantage, when adequately selected, of being human-readable and interpretable by experts. We quantify through a human evaluation that such features are not only discriminative, thus predictive for the classification, but also appear to be good candidates for providing explanatory elements of ADR mechanisms.
In conclusion, this work illustrates that simple models, fed with diverse and explicit knowledge sources such as those connected in the form of linked open data constitute an alternative to complex models, efficient but hard to interpret.
 A natural perspective is to combine such rich sources of background knowledge with models that are both highly performing (such as GCN) and interpretable.



\section*{Acknowledgements}
Authors acknowledge participants of the BioHackathon Paris 2018 where the seed of this work was planted, and in particular Miguel Boland and Patryk Jarnot. Authors also acknowledge anonymous reviewers of the Podium Abstract Session at MedInfo 2019 for their encouraging feedback on our positional abstract \cite{calvier:hal-02196134}. 

\section*{Funding}
The authors acknowledge the French National Research Agency (ANR) for funding PractiKPharma (ANR-15-CE23-0028) and FIGHT-HF (15-RHUS-0004) pro- jects.

\section*{Abbreviations}
ADR: Adverse drug reaction, AI: Artificial Intelligence, AUC: Area under the curve, usually under the ROC curve, DILI: Drug-induced liver injury,  GCN: Graph Convolutional Networks, GO: Gene ontology,  LOD: Linked open data, PGx: Pharmacogenomics, RAM: Random access memory, RDF: Resource description framework, SCAR: Severe cutaneous adverse reactions, SMILES: Simplified molecular-inputline-entry system, URI: Uniform resource identifier.

\section*{Availability of data and materials}
PGxLOD is available at \href{https://pgxlod.loria.fr/}{\small{https://pgxlod.loria.fr/}}. 
DILIRank is available at \href{https://www.fda.gov/science-research/liver-toxicity-knowledge-base-ltkb/drug-induced-liver-injury-rank-dilirank-dataset}{\small{https://www.fda.gov/science-research/liver-toxicity-knowledge-base-ltkb/drug-induced-liver-injury-rank-dilirank-dataset}}. RegiSCAR \\ ``Drug notoriety list" is available at  \href{http://www.regiscar.org/cht/pdf/Drug\%20Notoriety\%202015.\%20revised\%20may\%202017.xls}{\small{http:/  /www.regiscar.org/cht/pdf/Drug\%20Notoriety\%202015.\%20 revised\%20may\%202017.xls}}.
Rules and features generated and analysed during this study are included in the Additional information section.
Answers of the manual evaluation are included in the Additional file 1, available upon request.




\section*{Authors' contributions}
EB, PM, CB, MST and AC designed the study. EB, PM and FC mapped expert classifications to PGxLOD. EB and PM extracted the features, trained and evaluated the classifiers. EB, PM, MST and AC designed the manual evaluation. AC supervised it. CB, CNC and NP performed the manual evaluation of features. PM and AC were major contributors in writing the manuscript. EB and MST participated to the writing. All authors commented on the manuscript.
All authors read and approved the final manuscript.

\bibliographystyle{vancouver} 
\bibliography{bibliography}      

\newpage
\section*{Additional information}

 \subsection*{Additional file 1}
 This file include the results of the manual evaluation performed by three experts about the explanatory character of predictive features. Available on demand.
 
\newpage
\newgeometry{margin=1.2cm}
\begin{landscape}

\    

\vspace{4.5cm}

\begin{table}[!h]\begin{center}
\begin{tabular}{l|ccc}

      \textit{Name} & \textit{Rule} & \textit{Support} &  \textit{Nb of FP} \\
      \hline
      Rule1 & $\neg ( \xrightarrow{\text{involvedIn}} \text{Pathway} \xrightarrow{\text{contains}} \text{Compound} \xrightarrow{\text{interactsWith}} \textbf{Cytochrome P450 2A6}) \wedge \textbf{Cytochrome P450 3A4} \ \ \Rightarrow \ \ \oplus $ & 106.79 & 14.81\\
      \hline
      \multirow{3}*{Rule2} & $\neg ( \xrightarrow{\text{involvedIn}} \text{Pathway} \xrightarrow{\text{contains}} \text{Compound} \xrightarrow{\text{interactsWith}} \textbf{Cytochrome P450 2A6) } \wedge \textbf{Vitamin digestion and absorption } \wedge $ & \multirow{3}*{24.27} & \multirow{3}*{4.11}\\
      & $\neg ( \xrightarrow{\text{involvedIn}} \text{Pathway} \xrightarrow{\text{associatedWith}} \text{Disease} \xrightarrow{\text{interactsWith}} \textbf{Calcium signaling pathway}) \wedge $ & & \\ 
      & $\neg ( \xrightarrow{\text{targets}} \text{Protein} \xrightarrow{\text{cellularComponent}} \text{Integral component of plasma membrane})  \ \ \Rightarrow  \ \ \oplus $ & & \\
      \hline
      \multirow{3}*{Rule3} & $\neg ( \xrightarrow{\text{involvedIn}} \text{Resource} \xrightarrow{\text{contains}} \text{Drug} \xrightarrow{\text{contains}} \text{Phenols})  \wedge  $ & \multirow{3}*{37.98} & \multirow{3}*{11.52}\\
      & $\neg ( \xrightarrow{\text{binds}} \text{Resource} \xrightarrow{\text{associatedWith}} \text{Drug} \xrightarrow{\text{associatedWith}} \text{Resource) } \wedge $ & & \\ 
      & $\neg ( \xrightarrow{\text{associatedWith}} \text{Resource} \xrightarrow{\text{associatedWith}} \text{Resource)}  \ \ \Rightarrow  \ \ \oplus $ & & \\
      \hline
       \multirow{2}*{Rule4}& $ \xrightarrow{\text{interactsWith}} \text{Enzyme} \xrightarrow{\text{produces}} \text{Drug} \xrightarrow{\text{involvedIn}} \textbf{Biosynthesis of secondary metabolites } \wedge $ & \multirow{2}*{14.68} & \multirow{2}*{0.82} \\ 
      & $ \xrightarrow{\text{interactsWith}} \text{Enzyme} \xrightarrow{\text{cellularComponent}} \text{Endoplasmic reticulum }  \ \ \Rightarrow  \ \ \oplus $ & & \\
      \hline
      \multirow{2}*{Rule5}& $ \xrightarrow{\text{associatedWith}} \text{Drug} \xrightarrow{\text{associatedWith}} \text{Resource} \xrightarrow{\text{associatedWith}} \text{Resource } \wedge $ & \multirow{2}*{10.47} & \multirow{2}*{1.65} \\ 
      & $ \neg ( \xrightarrow{\text{interactsWith}} \text{Enzyme} \xrightarrow{\text{molecularFunction}} \text{Oxidoreductase activity})  \ \ \Rightarrow  \ \ \oplus $ & & \\
      \hline
      \multirow{2}*{Rule6}& $  \xrightarrow{\text{targets}} \text{Protein} \xrightarrow{\text{molecularFunction}} \text{Oxidoreductase activity } \wedge $ & \multirow{2}*{5.86} & \multirow{2}*{0.82} \\ 
      & $ \xrightarrow{\text{involvedIn}} \text{Resource} \xrightarrow{\text{contains}} \text{Organic amino compound} \xrightarrow{\text{interactsWith}} \textbf{Cytochrome P450 1A1 }  \ \ \Rightarrow  \ \ \oplus $ & & \\
\end{tabular}
\caption{\label{tab:s1} JRip rules learned from the DILI expert classification. Only positive rules (\textit{i.e.}, rules with $\oplus$ on the right-hand side) are generated. 
$Support$ of rules is the number of examples satisfying the rule. \textit{Nb of FP} is the number of false positives.
These two metrics are float because of weights associated with examples to balance classes in the training. Note that the second and third attributes of Rule2 and the first attribute of Rule 5 are path patterns that only contain general classes. We exclude this kind of attributes of our analysis because they are not sufficiently explanatory. In bold font are entities (final or not) of path patterns and neighbors.}
\end{center}
\end{table}

\end{landscape}
\restoregeometry
\newpage
\newgeometry{margin=1.2cm}
\begin{landscape}

\    

\vspace{4.5cm}

\begin{table}[!h]\begin{center}
\begin{tabular}{l|ccc}

      \textit{Name} & \textit{Rule} & \textit{Support} &  \textit{Nb of FP} \\
      \hline
      \multirow{2}*{Rule1} & $\neg ( \xrightarrow{\text{targets}} \text{Protein} \xrightarrow{\text{molecularFunction}} \text{Iron ion binding) } \wedge 
      \xrightarrow{\text{interactsWith}} \text{Enzyme} \xrightarrow{\text{biologicalProcess}}  \text{Electron transport } \wedge $ & \multirow{2}*{88.82} & \multirow{2}*{8.11}\\
      & $\neg ( \xrightarrow{\text{targets}} \text{Protein} \xrightarrow{\text{biologicalProcess}} \text{Positive regulation of cell proliferation }  \ \ \Rightarrow  \ \ \oplus $ & & \\
      \hline
      \multirow{2}*{Rule2} & $\neg ( \xrightarrow{\text{targets}} \text{Protein} \xrightarrow{\text{molecularFunction}} \text{Iron ion binding) } \wedge  $ & \multirow{2}*{34.12} & \multirow{2}*{3.38}\\
      & $\xrightarrow{\text{transportedBy}} \text{Protein} \xrightarrow{\text{cellularComponent}} \text{Membrane } \ \ \Rightarrow  \ \ \oplus $ & & \\
      \hline
      \multirow{2}*{Rule3} & $\xrightarrow{\text{targets}} \text{Protein} \xrightarrow{\text{molecularFunction}} \text{Nucleotide binding } \wedge \textbf{Cytochrome P450 2B6}$  & \multirow{2}*{42.70} & \multirow{2}*{8.11}\\
      & $\neg ( \xrightarrow{\text{interactsWith}} \text{Enzyme} \xrightarrow{\text{molecularFunction}} \text{Oxidoreductase activity})  \ \ \Rightarrow  \ \ \oplus $ & & \\ 
      \hline
       \multirow{3}*{Rule4}& $ \neg ( \xrightarrow{\text{involvedIn}} \text{Resource} \xrightarrow{\text{contains}} \text{Sulfonamide} \xrightarrow{\text{transportedBy}} \textbf{Solute carrier organic anion transporter family member 1A2) } \wedge $ & \multirow{3}*{52.70} & \multirow{3}*{19.60} \\ 
      & $ \neg (\xrightarrow{\text{interactsWith}} \textbf{Cytochrome P450 3A5) } \wedge 
      \neg ( \textbf{Calcium signaling pathway) }  \wedge $ & & \\
      
      & $ \neg ( \xrightarrow{\text{transportedBy}} \text{Protein} \xrightarrow{\text{molecularFunction}} \text{ATPase activity) }  \ \ \Rightarrow  \ \ \oplus $ & & \\
      \hline
      Rule5& $ \xrightarrow{\text{involvedIn}} \textbf{Calcium signaling pathway} \wedge \neg (  \xrightarrow{\text{involvedIn}} \textbf{Antifungal agents) }  \ \ \Rightarrow  \ \ \oplus $ & 8.36 & 0.68 \\ 
\end{tabular}
\caption{\label{tab:s2}  JRip rules learned from the SCAR expert classification. Only positive rules (\textit{i.e.}, rules with $\oplus$ on the right-hand side) are generated. 
$Support$ of rules is the number of examples satisfying the rule. \textit{Nb of FP} is the number of false positives.
These two metrics are float because of weights associated with examples to balance classes in the training. In bold font are entities (final or not) of path patterns and neighbors.}
\end{center}
\end{table}

\end{landscape}
\restoregeometry

\clearpage

\begin{table}
    \centering
    \caption{Features (neighbors and path patterns) associated with DILI extracted from JRip rules, and presented to experts for evaluating their explanatory potential with regard to DILI.}
    \begin{tabular}{lc}
        \hline
         1: & $\neg ( \xrightarrow{\text{involvedIn}} \text{Pathway} \xrightarrow{\text{contains}} \text{Compound} \xrightarrow{\text{interactsWith}} \textbf{Cytochrome P450 2A6})$ \\
         2: & $\textbf{Cytochrome P450 3A4}$ \\
         3: & $\textbf{Vitamin digestion and absorption}$ \\
         4: & $\neg ( \xrightarrow{\text{involvedIn}} \text{Pathway} \xrightarrow{\text{associatedWith}} \text{Disease} \xrightarrow{\text{interactsWith}} \textbf{Calcium signaling pathway})$ \\
         5: & $\neg ( \xrightarrow{\text{targets}} \text{Protein} \xrightarrow{\text{cellularComponent}} \text{Integral component of plasma membrane}$ \\
         6: & $\neg ( \xrightarrow{\text{involvedIn}} \text{Resource} \xrightarrow{\text{contains}} \text{Drug} \xrightarrow{\text{contains}} \text{Phenols})$ \\
         7: & $\xrightarrow{\text{interactsWith}} \text{Enzyme} \xrightarrow{\text{produces}} \text{Drug} \xrightarrow{\text{involvedIn}} \textbf{Biosynthesis of secondary metabolites}$ \\ 
         8: & $\xrightarrow{\text{interactsWith}} \text{Enzyme} \xrightarrow{\text{cellularComponent}} \text{Endoplasmic reticulum }$ \\
         9: & $\neg ( \xrightarrow{\text{interactsWith}} \text{Enzyme} \xrightarrow{\text{molecularFunction}} \text{Oxidoreductase activity})$ \\
         10 & $\neg ( \xrightarrow{\text{targets}} \text{Protein} \xrightarrow{\text{molecularFunction}} \text{Oxidoreductase activity}$ \\
         11: & $\xrightarrow{\text{involvedIn}} \text{Resource} \xrightarrow{\text{contains}} \text{Organic amino compound} \xrightarrow{\text{interactsWith}} \textbf{Cytochrome P450 1A1 } $\\

        \hline
    \end{tabular}
    \label{tab:s3}
\end{table}

\begin{table}
    \centering
    \caption{Features (neighbors and path patterns) associated with SCAR extracted from JRip rules, and presented to experts for evaluating their explanatory potential with regard to SCAR.}
    \begin{tabular}{lc}
        \hline
         1: & $\neg (\xrightarrow{\text{targets}} \text{Protein} \xrightarrow{\text{molecularFunction}} \text{Iron ion binding})$ \\
         2: & $\xrightarrow{\text{interactsWith}} \text{Enzyme} \xrightarrow{\text{biologicalProcess}}  \text{Electron transport}$ \\
         3: & $\neg (\xrightarrow{\text{targets}} \text{Protein} \xrightarrow{\text{biologicalProcess}} \text{Positive regulation of cell proliferation}$ \\
         4: & $\xrightarrow{\text{transportedBy}} \text{Protein} \xrightarrow{\text{cellularComponent}} \text{Membrane}$ \\
         5: & $\xrightarrow{\text{targets}} \text{Protein} \xrightarrow{\text{molecularFunction}} \text{Nucleotide binding} $ \\
         6: & $\textbf{Cytochrome P450 2B6}$ \\
         7: & $\neg (\xrightarrow{\text{interactsWith}} \text{Enzyme} \xrightarrow{\text{molecularFunction}} \text{Oxidoreductase activity})$ \\ 
         \multirow{2}*{8:} & $\neg ( \xrightarrow{\text{involvedIn}} \text{Resource} \xrightarrow{\text{contains}} \text{Sulfonamide} \xrightarrow{\text{transportedBy}} \textbf{Solute } $ \\
          & $ \textbf{carrier organic anion transporter family member 1A2})$\\
         9: & $\neg (\xrightarrow{\text{interactsWith}} \textbf{Cytochrome P450 3A5})$ \\
         10 & $\neg ( \textbf{Calcium signaling pathway})$ \\
         11: & $\neg (\xrightarrow{\text{transportedBy}} \text{Protein} \xrightarrow{\text{molecularFunction}} \text{ATPase activity})$ \\
         12: & $\xrightarrow{\text{involvedIn}}$\textbf{Calcium signaling pathway}\\
         13: & $\neg (\xrightarrow{\text{involvedIn}}  \textbf{Antifungal agents})$\\

        \hline
    \end{tabular}
    \label{tab:s4}
\end{table}

\end{document}